%% file: arxiv.tex
\documentclass[10pt, a4paper]{article}

\usepackage[final]{lrec2026} 

\usepackage{booktabs}
\usepackage{url}
\usepackage{soul}
\usepackage{tabularx}
\usepackage{multirow}
\usepackage{enumitem}

\usepackage{listings}
\usepackage{xcolor}
\usepackage{subcaption}
\usepackage{comment}
\usepackage{graphicx}

\definecolor{codegreen}{rgb}{0,0.6,0}
\definecolor{codegray}{rgb}{0.5,0.5,0.5}
\definecolor{codepurple}{rgb}{0.58,0,0.82}
\definecolor{backcolour}{rgb}{0.95,0.95,0.92}

\lstdefinestyle{mystyle}{
    backgroundcolor=\color{backcolour},   
    commentstyle=\color{codegreen},
    keywordstyle=\color{magenta},
    numberstyle=\tiny\color{codegray},
    stringstyle=\color{codepurple},
    basicstyle=\ttfamily\footnotesize,
    breakatwhitespace=false,         
    breaklines=true,                 
    captionpos=b,                    
    keepspaces=true,                 
    numbers=left,                    
    numbersep=5pt,                  
    showspaces=false,                
    showstringspaces=false,
    showtabs=false,                  
    tabsize=2
}

\usepackage{mdframed}
\usepackage{amsmath}

\definecolor{beige}{rgb}{0.96, 0.96, 0.86}

\usepackage{tcolorbox} 
\usepackage{xcolor}
\definecolor{steel}{RGB}{70, 130, 180}

\title{TempPerturb-Eval: On the Joint Effects of Internal Temperature and External Perturbations in RAG Robustness}

\name{Yongxin Zhou, Philippe Mulhem, Didier Schwab} 

\address{Univ. Grenoble Alpes, CNRS, Grenoble INP, LIG, 
         38000, Grenoble, France \\
         \{yongxin.zhou, philippe.mulhem, didier.schwab\}@univ-grenoble-alpes.fr\\}

\abstract{
The evaluation of Retrieval-Augmented Generation (RAG) systems typically examines retrieval quality and generation parameters like temperature in isolation, overlooking their interaction. This work presents a systematic investigation of how text perturbations (simulating noisy retrieval) interact with temperature settings across multiple LLM runs. We propose a comprehensive RAG Perturbation-Temperature Analysis Framework that subjects retrieved documents to three distinct perturbation types across varying temperature settings. Through extensive experiments on HotpotQA with both open-source and proprietary LLMs, we demonstrate that performance degradation follows distinct patterns: high-temperature settings consistently amplify vulnerability to perturbations, while certain perturbation types exhibit non-linear sensitivity across the temperature range. Our work yields three key contributions: (1) a diagnostic benchmark for assessing RAG robustness, (2) an analytical framework for quantifying perturbation-temperature interactions, and (3) practical guidelines for model selection and parameter tuning under noisy retrieval conditions.
 \\ \newline  
}

\begin{document}

\maketitleabstract

\section{Introduction}

Retrieval Augmented Generation (RAG)~\citep{NEURIPS2020_6b493230} is a prompt engineering strategy that augments the internal capacity of Large Language Models (LLMs) with external knowledge. 
In a RAG, incorrect retrieved documents can introduce external noise that affects output quality \citep{fang-etal-2024-enhancing, wang-etal-2025-astute, kang2025getalpautomin2025leveragingrag}. One RAG output also depends on the hyperparameters of its LLM, e.g., the \textit{temperature}: the generated text is more (resp. less) deterministic for low (resp. large) temperature values~\citep{DBLPHoltzmanBDFC20}.

Furthermore, \textit{Perturbations} serve as adversarial examples in evaluating RAG robustness, simulating small input changes that can deceive models into incorrect predictions. These modifications help quantify how much specific input features must change to alter model outcomes~\citep{anand2022explainableinformationretrievalsurvey}. Previous research has employed various perturbation strategies for RAG question-answering systems, such as the \textit{leave-one-token-out} approach that systematically removes individual sentences from input texts~\citep{10.1145/3626772.3657660}.

However, existing literature overlooks a critical dimension: the interaction between perturbations and generation hyperparameters, particularly temperature. This gap is significant given that temperature substantially affects output quality across various tasks~\citep{renze-2024-effect,du2025optimizing, LI2025242}, with low temperature values not always constituting the optimal choice. Consequently, current perturbation-based evaluations may yield misleading robustness assessments by failing to account for temperature variability.

Our work addresses this limitation by systematically investigating how perturbations interact with temperature settings in RAG systems. While traditional evaluations examine retrieval quality and generation parameters in isolation, they overlook their practical interdependence. By integrating both dimensions, our framework provides more reliable faithfulness explanations and accurate robustness measurements under realistic deployment conditions.

\begin{figure*}[!ht]
\begin{center}
    \includegraphics[trim={1.3cm 3.8cm 1.3cm 3.8cm}, clip, width=\textwidth]{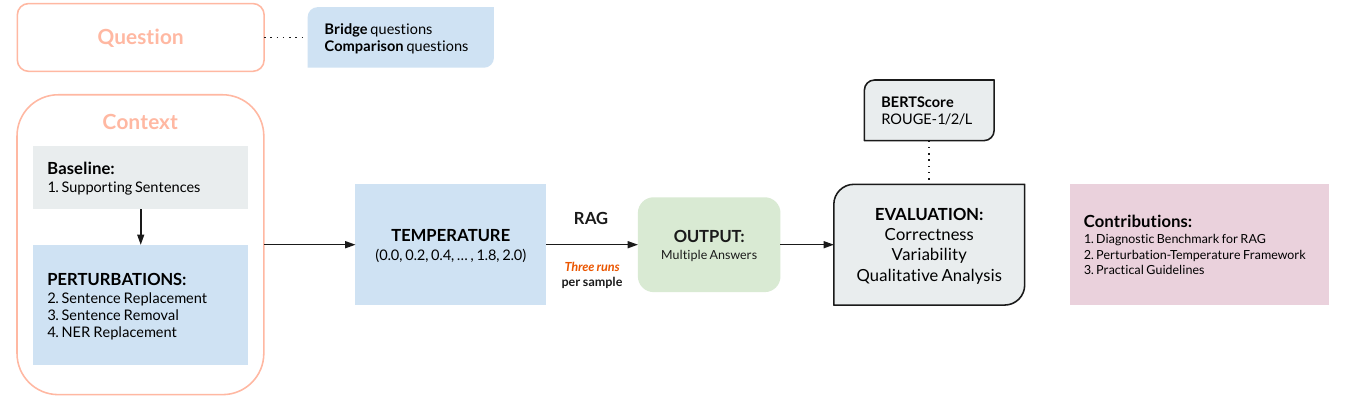}
\caption{RAG Perturbation-Temperature Analysis Framework. The methodology stresses system robustness along two axes: external context perturbations (replacement, removal, NER substitution) and internal LLM temperature variation. The evaluation measures correctness and output variability across these conditions to establish a benchmark and derive practical guidelines.}
\label{fig:framework}
\end{center}
\end{figure*}

We approach the RAG LLM as a black box and experimentally quantify the \textit{temperature effect} using the HotpotQA dataset~\citeplanguageresource{yang-etal-2018-hotpotqa}, a benchmark for multi-hop question answering that demands complex reasoning and yields explanation-rich answers. We pair this dataset with systematic text perturbations to simulate noisy retrieval. To our knowledge, this represents the first comprehensive study of temperature-perturbation interactions in RAG systems\footnote{The source code and experimental framework are available at \url{https://github.com/yongxin2020/TempPerturb-Eval}, and the complete dataset of model inputs and generations is released on Hugging Face at \url{https://huggingface.co/datasets/yongxin2020/TempPerturb-Eval-data}.}.
As shown in Figure~\ref{fig:framework}, our work systematically investigates this complex interplay, with three key contributions:

\begin{itemize}[noitemsep,topsep=0pt,parsep=0pt,partopsep=0pt]
    \item A comprehensive taxonomy of perturbations for RAG, synthesizing and categorizing methods from information retrieval literature.
    \item A publicly released diagnostic benchmark that quantifies RAG robustness across 440 experimental conditions, spanning multiple models, temperatures, perturbation types, and question types.
    \item An analytical framework that models the joint impact of temperature and perturbations, accompanied by practical guidelines for robust model deployment.
\end{itemize}
 
\section{Related Work}
\label{sec:related_work}

\subsection{Perturbations in IR and RAG}

Previous work have used perturbations as adversarial examples to examine the robustness of Information Retrieval (IR) models \citep{raval2020wordtimeadversarialattacks, 10.1145/3576923, liu2024robustneuralinformationretrieval}.
Typical perturbations include removing, adding, or replacing words, phrases, sentences, passages, or entire documents.
For instance, ~\citet{raval2020wordtimeadversarialattacks} found that even minimal token changes (1-3), an \textit{attacker} can produce semantically similar perturbed documents capable of fooling document rankers. 

Perturbations are also used in counterfactual explanations in explainable IR, where the closest samples on which the model makes a different prediction serve as example-based explanations~\citep{10.1007/978-3-031-44067-0_2}.
Some demonstrations \citep{Rorseth2023CREDENCECE, rorseth2024ragemachineretrievalaugmentedllm} provide examples of explanations for RAGs obtained using different perturbation methods, but they do not evaluate their proposals.
In the RAG context specifically, only the \textit{leave-one-token-out} strategy has been evaluated on a question-answering task~\citep{10.1145/3626772.3657660}. 

We review the relevant literature to provide an overview of perturbation methods from explainable information retrieval that can assess the robustness of IR and RAG systems \citep{zhou:hal-05324909}. Table \ref{tab:perturbations} summarizes these approaches, organized according to the following dimensions:

\begin{itemize}
    \item \textbf{Target}: The IR component being perturbed (document or query).
    \item \textbf{Perturbation Category}: The high-level strategy for modifying content (e.g., subset selection, addition, replacement).
    \item \textbf{Specific Method}: The concrete technique used to implement the perturbation.
    \item \textbf{Granularity}: The textual unit affected by the perturbation (e.g., token, sentence, passage).
    \item \textbf{Description and Application}: An explanation of the method and its use cases.
\end{itemize}

\begin{table*}[!htb]
    \scriptsize
    \centering
    \resizebox{\textwidth}{!}{
    \input{tables/perturbations_taxonomy}
    }
    \caption{Taxonomy of perturbation methods for evaluating Information Retrieval (IR) and Retrieval-Augmented Generation (RAG) systems. 
    }
    \label{tab:perturbations}
\end{table*}

The retrieval perturbations summarized in Table~\ref{tab:perturbations} introduce controlled deviations from an ideal retrieval result. The most direct way to link these perturbations to Information Retrieval (IR) performance is through document- or passage-level modifications. For example, removing a relevant document from the ranked list demonstrably lowers standard evaluation metrics such as precision, recall, and nDCG. A similar effect is observed when swapping the positions of a relevant and a non-relevant passage.

In contrast, perturbations at a finer granularity, such as synonym replacement, present a more complex scenario, as they may not inherently alter a document's underlying relevance. Other token-level perturbations, including random noise injection or entity replacement, are even more challenging to evaluate using classical IR metrics, which assume relevance judgments are based on unmodified text.
To address this, the relevance of a perturbed document can be estimated by computing its similarity to the original version and applying a threshold, providing a pseudo-relevance score. Alternatively, LLM-as-a-judge methodologies \citep{gu2025surveyllmasajudge} offer a flexible approach. These techniques effectively augment original relevance assessments, analogous to data augmentation strategies used in computer vision \citep{szegedy2014goingdeeperconvolutions}, enabling a more comprehensive evaluation of robustness under textual variations.

\subsection{LLM Temperature Impact}

LLMs generate token sequences using token logits $l_k$ for each token $v_k$. The temperature modifies the output probabilities of the tokens so that the distribution peaks (resp. is flat) for large (resp. low) temperature values. Then it also influences the sampling of these tokens and therefore the whole generation. High-temperature values are supposed to add diversity to generation: several \textit{runs} of the same prompt may generate very different responses. With the notation of~\citet{renze-2024-effect}, the probability of $v_k$, using the temperature hyperparameter $T$, is: 

\begin{eqnarray}
p(v_{k}) = \frac{e^{l_{k}/T}}{ \sum\limits_{i} e^{l_{i}/T}}
\end{eqnarray}

\textit{OpenAI} and \textit{DeepSeek} API documentations provide temperature recommendations for several tasks without documented support for these values.
However, ~\citet{renze-2024-effect} found that varying sampling temperature from $0.0$ to $1.0$ does not yield statistically significant differences in problem-solving performance on multiple-choice question answering (MCQA) tasks across several LLMs. 
Building on this insight, we extend the analysis to RAG systems by investigating how perturbations interact with temperature.

\section{Methodology}
\label{sec:methodology}

We seek to estimate the impact, if any, of the LLM temperature hyperparameter when perturbing a RAG LLM input. We cope with the internal variability coming from the LLM by presenting the same prompt (perturbed or non-perturbed) several times. Using this, our methodology assesses the behavior of the perturbations along the temperature evolution.
We compare each generated text by the LLM with a processed ground-truth (see Section~\ref{ssec:eval}) using classical semantic similarity measures and compute the mean, variance and standard deviation for the same prompt. We then build graphics that present these comparisons.

This analysis allows us to determine: (i) whether certain perturbation types consistently degrade performance across all temperature values; (ii) whether the effect of specific perturbations at certain temperatures is statistically indistinguishable from the non-perturbed baseline; and (iii) whether the relative impact of different perturbations changes with temperature (e.g., if \textit{Perturbation A} has more impact than \textit{Perturbation B} at a low temperature, but less impact at a high temperature).

\section{Experiments}
\label{sec:experiments_design}

\subsection{Dataset and Perturbations}

We selected for our experiments the HotpotQA~\citeplanguageresource{yang-etal-2018-hotpotqa} dataset, dedicated to question-answering (QA) systems that perform complex reasoning and provide explanations for their answers. It contains 113k Wikipedia-based QA pairs\footnote{\url{https://huggingface.co/datasets/hotpotqa/hotpot_qa}}. 
This dataset was selected for three key characteristics: (i) sentence-level supporting facts that facilitate clean baseline establishment; (ii) multi-hop QA structure requiring reasoning across multiple documents, making it well-suited for perturbation testing; and (iii) availability of ground-truth answers for each query. 

Furthermore, the classification of queries into ``bridge'' and ``comparison'' types enables the investigation of system behavior across distinct reasoning categories.
\textit{Bridge} questions are those where, to arrive at the answer, one must first identify a bridge entity and then find the answer in relation to it. The other type of multi-hop questions consists of \textit{Comparison} questions, which require comparing two entities from the same category. A subset of these comparison questions are yes/no questions.

More precisely, we utilized the training set of the \textit{fullwiki} version of the HotpotQA dataset. After analyzing the statistics of the data set, we randomly selected 100 samples for each category of facts (2, 3 and 4 facts) and for each type of question (``bridge'' and ``comparison''), resulting in a total of 600 samples for experimentation\footnote{This sample size balances statistical reliability against computational constraints, given our fine-grained experimental design of 3 runs per (model, temperature, perturbation, query) condition.}.

In our experiments, we establish a baseline using all original supporting sentences. Building upon this baseline, we systematically introduce three types of perturbations, selected for their relevance to real-world retrieval errors and alignment with established evaluation frameworks such as RAG-Ex \citep{10.1145/3626772.3657660}.
Our perturbation strategy includes\footnote{For each sample, the number of altered sentences was scaled by fact count: one sentence for 2-fact samples, one (33\%) for 3-fact, and two for 4-fact samples. All other supporting sentences remained unperturbed. 
}: (1) \textbf{Sentence Replacement:} replacing the latter portion of supporting sentences with irrelevant sentences from the same title, which simulates retrieval of correct entities with incorrect evidence, a common and realistic failure mode in QA systems; (2) \textbf{Sentence Removal:} deleting the latter half of supporting sentences; and (3) \textbf{NER Replacement:} masking named entities in the last supporting sentence(s) by replacing them with \textit{[MASK]} tokens, focusing particularly on title-related entities to probe model sensitivity. 

This procedure generated three perturbed input conditions in addition to the original baseline. The resulting setup allows for a controlled investigation of core perturbation effects against a stable reference point, establishing a reproducible framework for future robustness studies.

\subsection{Models and RAG Configuration}
We conducted experiments with five LLMs, categorized as follows: 

\begin{itemize}[noitemsep,topsep=0pt,parsep=0pt,partopsep=0pt]
    \item \textbf{Proprietary GPT Models\footnote{\url{https://platform.openai.com/docs/models}}:} \texttt{gpt-3.5-turbo}, \texttt{gpt-4o};
    \item \textbf{Open-Source LLaMA Models:} \texttt{Llama-3.1-8B-Instruct}\footnote{\url{https://huggingface.co/meta-llama/Llama-3.1-8B-Instruct}, pretrained and fine-tuned text models in 8B sizes.} and \texttt{Llama-3.2-1B-Instruct}\footnote{\url{https://huggingface.co/meta-llama/Llama-3.2-1B-Instruct}, pretrained and instruction-tuned generative models in 1B sizes.}.
    \item \textbf{DeepSeek reasoning model}: \texttt{deepseek-reasoner}\footnote{\url{https://api-docs.deepseek.com/guides/reasoning_model}. Before delivering the final answer, the model first generates a Chain of Thought (CoT) to enhance the accuracy of its responses.}. 
\end{itemize}

The chosen models (GPT-family, Llama-family, and deepseek-reasoner) offer a strategically diverse mix of architectural families, parameter scales, and commercial vs. open-weight availability, allowing us to evaluate robustness across different model types.

For each condition (model, temperature, perturbation type), we executed the same query three times to account for intrinsic stochasticity and to help distinguish the effect of the model’s internal noise (due to temperature) from that of external perturbations. All other LLM hyperparameters were set to their default values, except for max\_tokens, which is set to 1000. 

\subsection{Evaluation Methodology}
\label{ssec:eval}

\textbf{Evaluation Metrics.} 
While Exact Match (EM) and F1 are widely adopted metrics, their limitations in evaluating long-form generative outputs are well-documented. In RAG settings, models frequently produce elaborated answers containing correct core information alongside supplementary explanations. Consequently, EM scores may be artificially low despite semantic correctness, and these standard metrics often fail to capture subtle differences in perturbed answers. The F1 metric offers greater robustness by rewarding token-level overlap, but unlike binary or multiple-choice QA, real-world RAG systems generate free-form answers requiring more nuanced evaluation.

Therefore, we report similarity metrics, which better reflect nuanced changes than exact matching. For instance, BERTScore \citep{Zhang*2020BERTScore:} can detect minor perturbations, such as passive/active voice shifts or small rewrites that retain the same meaning, while being more sensitive to semantic alterations than token-based metrics. 

\textbf{Reference Answer Processing.}
The reference answers of HotpotQA are short, for example, ``flew in space” for the \textit{bridge} question type, and ``Yes” or ``No” for the \textit{comparison} question type. Since our framework uses a similarity measure to assess the influence of temperature and perturbations on the output, we transform the reference information into sentence form by combining the original question and the short answer. 
This is achieved using \texttt{GPT-4o} (with default hyperparameters) as the backbone model. The model is prompted with a combination of the question and the candidate answer, using the following template:

\begin{table}[htbp]
\centering
\begin{tcolorbox}[
  colback=steel!5,  
  colframe=steel!75, 
  title=Prompt for Complete Answer Generator,
  fonttitle=\bfseries,
  ]

\textbf{Question:} \texttt{\{question\}} \\
\textbf{Answer:} \texttt{\{answer\}} \\
Generate a complete and coherent answer based on the given question and answer, being as brief as possible:

\end{tcolorbox}
\end{table}

Because the generated output are lengthy, we extract only the first sentence --- or the first two if they began with ``Yes” or ``No” --- as reference answers for comparison.

Given that \texttt{GPT-4o}'s role was limited to the low-complexity task of formatting answers (e.g., converting ``flew in space” to ``Both X and Y are astronauts who flew in space”), we observed high formatting accuracy. To verify this, we manually inspected a subset of the generated answers, including the final reference answers used for evaluation. A spot-check of 20 samples revealed no hallucinations. 

\textbf{Metric Selection and Reporting.}
We evaluated semantic similarity using both BERTScore~\citep{Zhang*2020BERTScore:} and ROUGE-1/2/L \citep{lin-2004-rouge}, two standard metrics for natural language generation \citep{zhou-etal-2025-gpt}. While all metrics exhibited consistent trends across experimental conditions, we selected BERTScore as our primary measure due to its stronger alignment with human judgment of semantic equivalence. We report BERTScore F1 scores computed using the default \texttt{RoBERTa-large} model \citep{liu2019robertarobustlyoptimizedbert} as the backbone.

\section{Experimental Results and Analysis}
\label{sec:results_analysis}

\subsection{Correctness Analysis}
\label{ssec:ref_results_analysis}

\begin{figure*}[!htb]
    \centering
    \includegraphics[width=\linewidth, trim=6cm 0cm 6cm 0cm, clip]{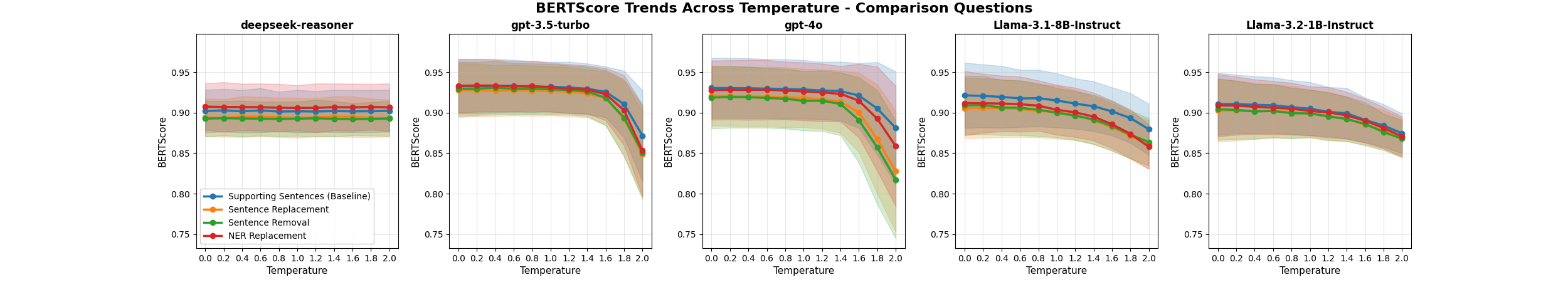}
    \includegraphics[width=\linewidth, trim=6cm 0cm 6cm 0cm, clip]{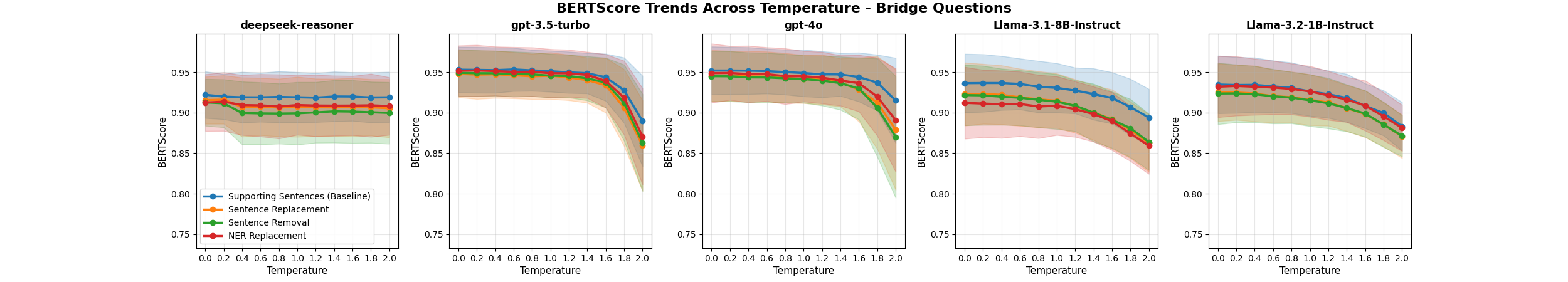}
    \caption{BERTScore trends across temperature variations for different models, comparing response types under perturbation. Solid lines represent mean scores across samples, while shaded areas denote $\pm$ standard deviation. The top row presents results for comparison questions; the bottom row presents results for bridge questions.}
    \label{fig:temperature_trends_bertscore}
\end{figure*}

We analyze BERTScore trends across temperature settings for different models in Figure~\ref{fig:temperature_trends_bertscore}. For each experimental condition, we compute the mean and standard deviation of scores across three runs, then aggregate these values across all samples per condition.

Our results reveal distinct model-specific temperature sensitivity patterns. While \texttt{deepseek-reasoner} maintains nearly invariant performance across the temperature range, GPT models exhibit degradation beginning at $T=1.4$. In contrast, Llama models demonstrate earlier performance deterioration at $T=0.6$, though with a more gradual decline slope compared to GPT models' sharper descent.

Taking \texttt{gpt-4o} for example (third column graphics from Figure~\ref{fig:temperature_trends_bertscore}), its results reveal that different perturbation types exhibit varying sensitivity to temperature increases. \textit{NER Replacement} induces minimal degradation at $T=2.0$, whereas \textit{Sentence Replacement} and \textit{Sentence Removal} lead to more substantial performance loss. Notably, all perturbation types demonstrate amplified sensitivity compared to baseline conditions as temperature rises, suggesting that temperature acts as a performance degradation amplifier. 

As temperature varies, we observe a shifting performance hierarchy. At lower temperatures ($T<1.4$), GPT models achieve the highest BERTScore, followed by Llama models and then \texttt{deepseek-reasoner}. However, this ranking reverses at higher temperatures ($T=2.0$), where \texttt{deepseek-reasoner} maintains consistent performance while GPT models degrade below the levels held by Llama models.

Question type (comparison vs. bridge) has minimal impact on temperature sensitivity, with both types exhibiting nearly identical degradation curves across all models and perturbations. This suggests that temperature effects are largely orthogonal to question complexity. However, absolute performance is consistently higher for bridge questions than for comparison questions across all models.

For deployment scenarios requiring temperature tuning, we recommend: (1) \texttt{deepseek-reasoner} for applications requiring consistent performance across diverse temperature settings; (2) GPT models with temperature ceilings of $T \leq 1.4$ to avoid sharp performance cliffs; and (3) Llama models with conservative temperature limits of $T \leq 0.6$ to maintain acceptable correctness levels.

\subsection{Output Variability Analysis}

\begin{figure*}[!htb]
    \centering
    \includegraphics[width=\linewidth]{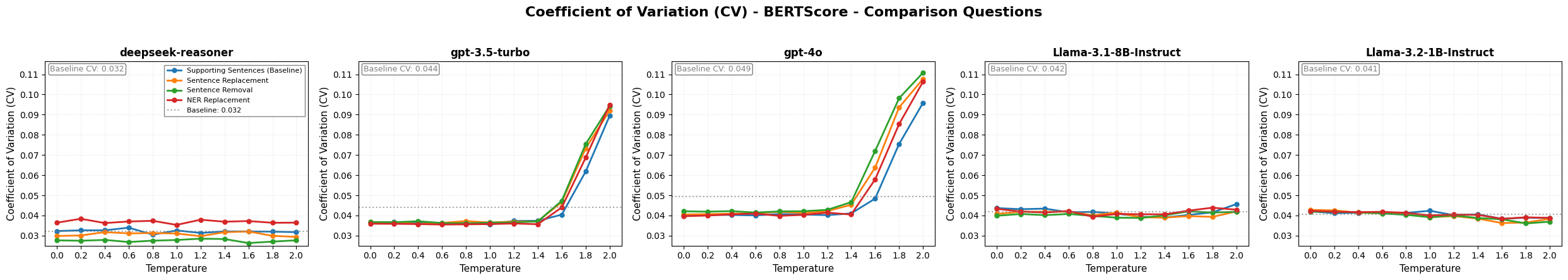}
    \includegraphics[width=\linewidth]{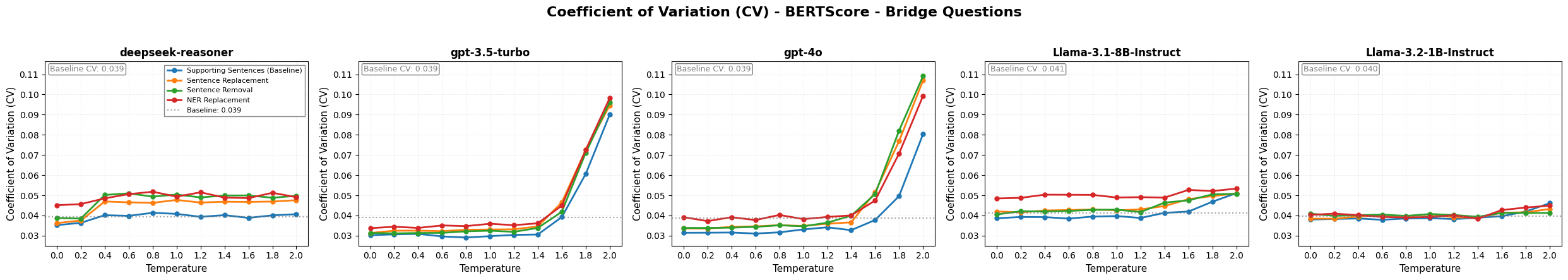}
    \caption{Coefficient of Variation (CV) for BERTScore across models, temperatures, and perturbation types. Each subplot displays CV trends for a model. The baseline CV value (average CV for the original, unperturbed context across all temperatures) is indicated in the top left of each subplot. The top row presents results for comparison questions; the bottom row presents results for bridge questions.}
    \label{fig:variability_bertscore}
\end{figure*}

To quantify performance sensitivity, we employ the Coefficient of Variation (CV), which measures relative variability by normalizing the standard deviation against the mean performance. Figure~\ref{fig:variability_bertscore} visualizes these results, with gray dotted lines indicating each model's stability baseline, calculated as the average CV for the original (unperturbed) context across all temperatures.

Our analysis reveals that temperature exerts a stronger influence on output variability than perturbation types in most models. However, Llama models exhibit distinct behavior: \texttt{Llama-3.2-1B-Instruct} shows no noticeable variations for comparison questions and bridge questions, whereas \texttt{Llama-3.1-8B-Instruct} exhibits variation that depends on both the perturbation type and the temperature for bridge questions.
GPT models demonstrate particularly high temperature sensitivity, with significant variability emerging at $T\geq1.4$. In contrast, \texttt{deepseek-reasoner} and Llama models maintain more consistent performance across the temperature range.
For the \texttt{deepseek-reasoner} model, \textit{NER Replacement} perturbations have the greatest impact on comparison questions, while all three perturbation types impact bridge questions, with sensitivity emerging from $T\geq0.2$.

The impact of question type on output variability is model- and perturbation-dependent. While bridge questions consistently show lowest variability in the unperturbed baseline across most models, comparison questions exhibit no consistent pattern: for instance, \textit{Sentence Removal} yields the lowest CV for \texttt{deepseek-reasoner}, but not for other models. This heterogeneity highlights the complex interplay between question type, perturbation, and model architecture.

\section{Qualitative Analysis of Model Sensitivity}

To complement our quantitative findings, we conducted a qualitative analysis of model behavior under varying temperatures and input perturbations. We selected \texttt{gpt-4o} and \texttt{deepseek-reasoner} for this analysis based on their contrasting sensitivity profiles observed in previous experiments: with \texttt{gpt-4o} demonstrating higher temperature sensitivity and \texttt{deepseek-reasoner} showing greater stability. We examined model outputs at two temperature extremes: $T=0.6$ (representing more deterministic generation) and $T=2.0$ (producing more stochastic outputs).

\subsection{BERTScore distributions}
\label{subsec:bertscore_distribution}

\begin{figure}[!ht]
    \centering
    \includegraphics[width=.48\columnwidth, trim=1.6cm 3cm 1.6cm 3cm, clip]{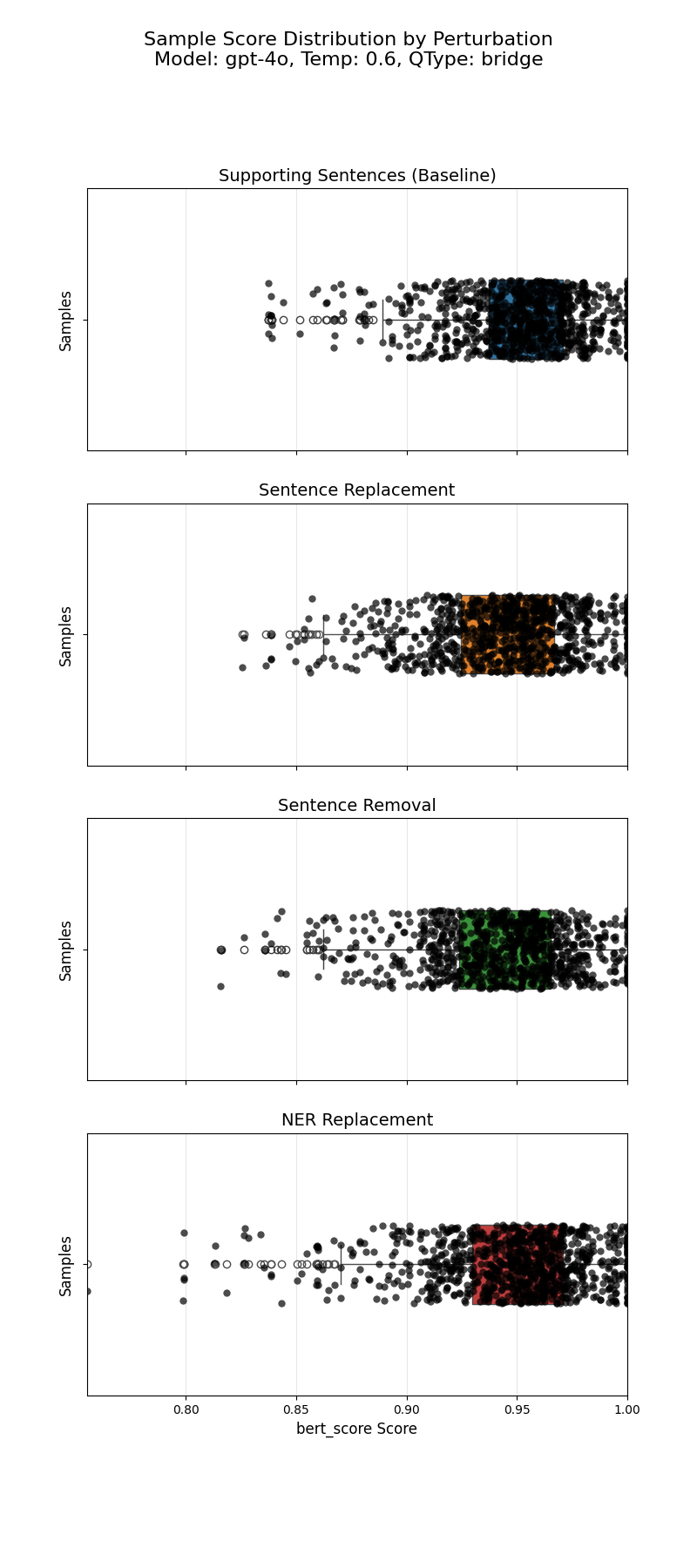}
    \includegraphics[width=.48\columnwidth, trim=1.6cm 3cm 1.6cm 3cm, clip]{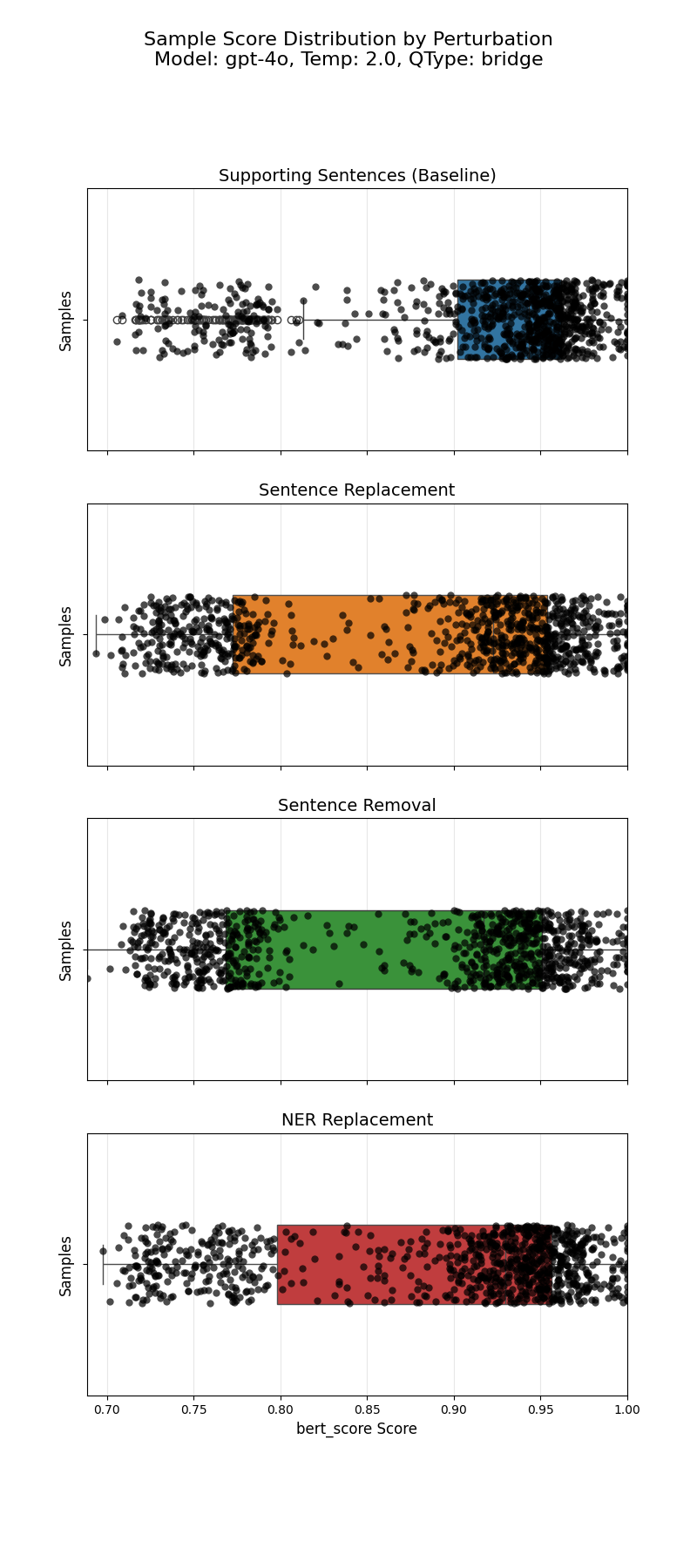}
    \caption{BERTScore distribution for \texttt{gpt-4o} on bridge questions across perturbation types at two temperatures (Left: $T=0.6$, Right: $T=2.0$). Each subplot shows a boxplot representing median, interquartile range, and whiskers, with individual sample scores (black dots) and outliers (white dots).} 
    \label{fig:sample_score_distribution_gpt_4o}
\end{figure}

\begin{figure}[!ht]
    \centering
    \includegraphics[width=.48\columnwidth, trim=1.6cm 3cm 1.6cm 3cm, clip]{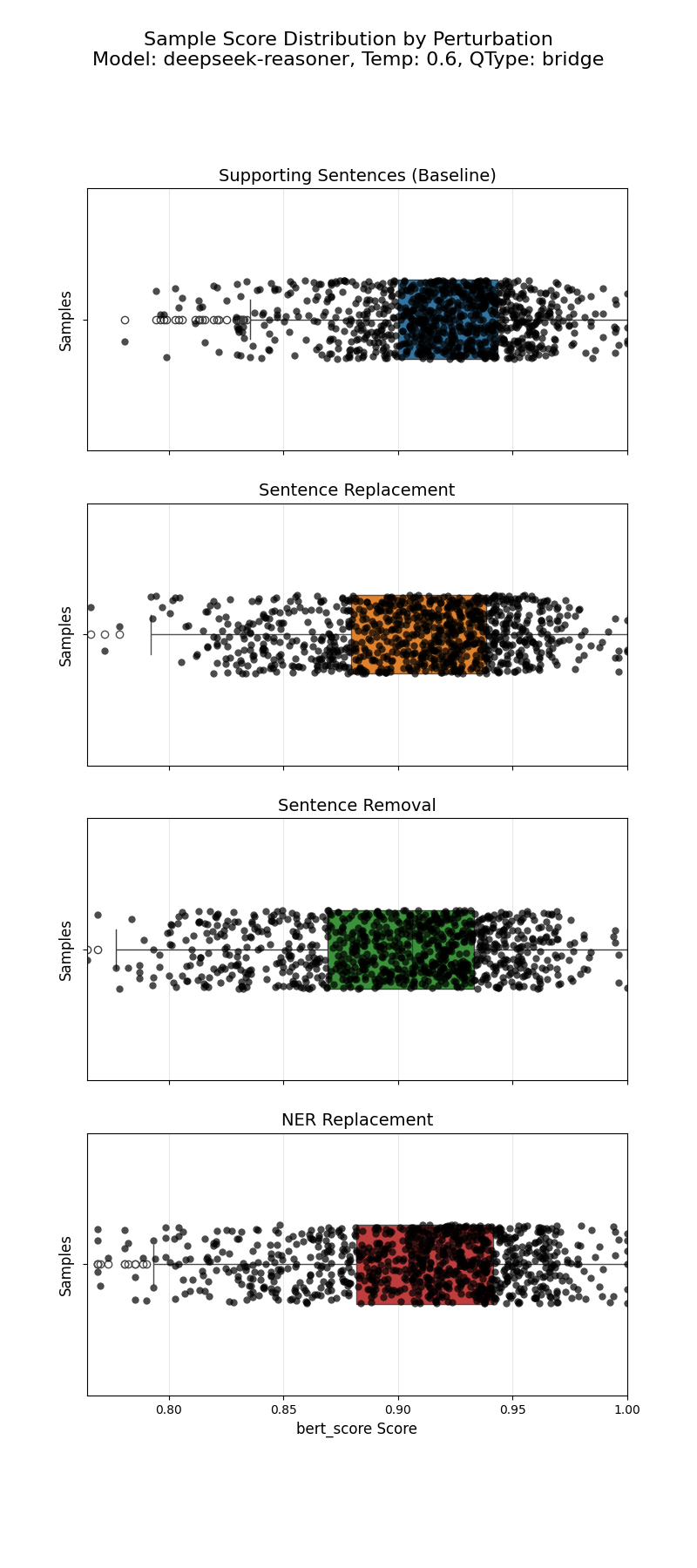}
    \includegraphics[width=.48\columnwidth, trim=1.6cm 3cm 1.6cm 3cm, clip]{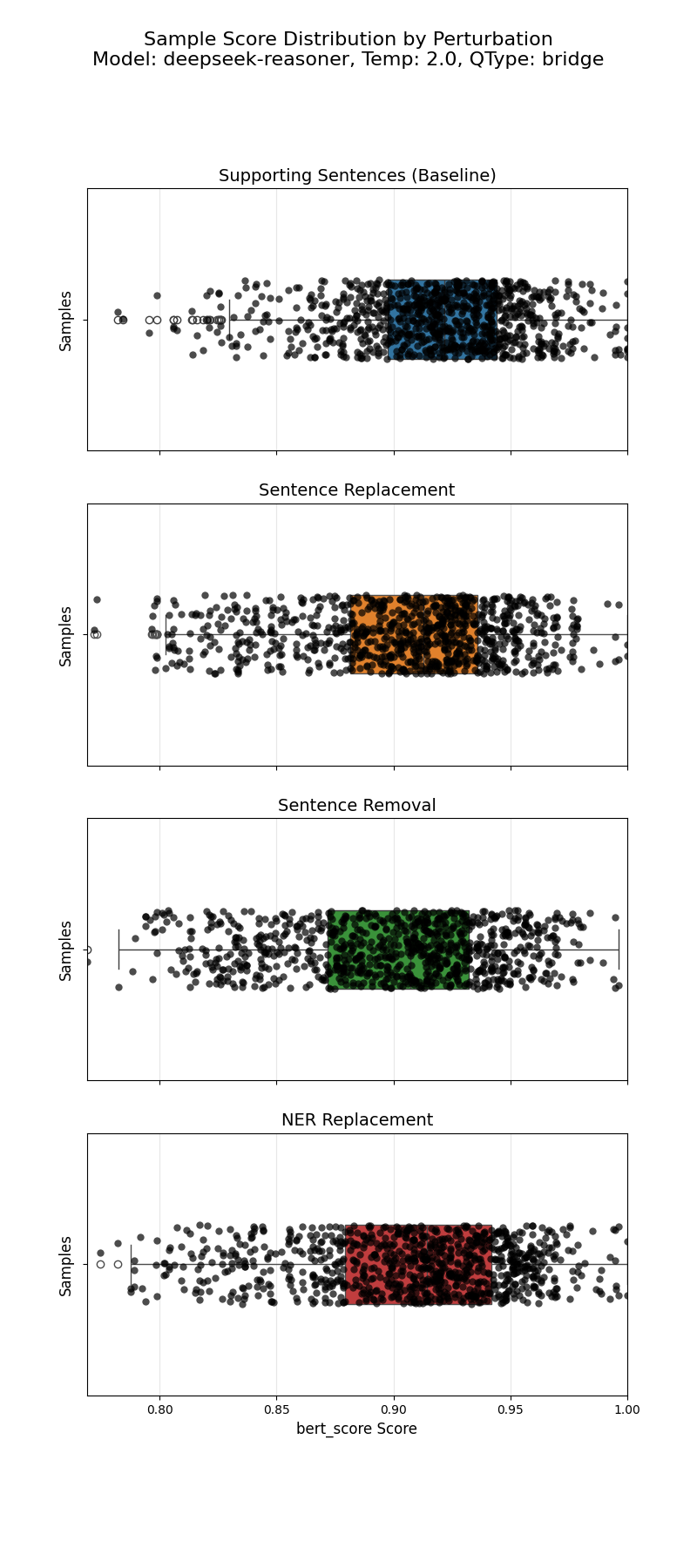}
    \caption{BERTScore distribution for \texttt{deepseek-reasoner} on bridge questions across perturbation types at two temperatures (Left: $T=0.6$, Right: $T=2.0$). Each subplot shows a boxplot representing the same elements as in Fig.~\ref{fig:sample_score_distribution_gpt_4o}.}
    \label{fig:sample_score_distribution_deepseek}
\end{figure}

Figures~\ref{fig:sample_score_distribution_gpt_4o} and~\ref{fig:sample_score_distribution_deepseek} illustrate the BERTScore distributions for bridge-type questions under different perturbations. Temperature significantly impacts output quality, particularly for \texttt{gpt-4o}. At $T=2.0$, performance degrades across all perturbations, with BERTScore values frequently falling between 0.70 and 0.80 and occasionally dropping below 0.70, indicating increased output variability and reduced semantic faithfulness at higher temperatures.

In contrast, \texttt{deepseek-reasoner} exhibits stability across temperature settings. While $T=2.0$ introduces slightly higher score variance, the median BERTScore remains consistent across temperatures for each perturbation type, suggesting more robust generation under temperature variation.

\subsection{Sample Analysis}

To identify representative cases of model sensitivity, we selected, for each model studied in Section \ref{subsec:bertscore_distribution}, temperature, question type, and perturbation type, the sample with the largest BERTScore gap between original and perturbed conditions. This method highlights key fragility patterns.

\textbf{Perturbation-Type Analysis.}
Our examination of these cases reveals distinct failure modes across perturbation types. \textit{Sentence Replacement} and \textit{Sentence Removal} perturbations frequently trigger model refusal behaviors, with responses such as ``The retrieved document does not provide...'' becoming common\footnote{For example, \texttt{gpt-4o} with \textit{Sentence Replacement} at $T=0.6$ output: ``The retrieved document does not provide specific information about the campus sizes of Indiana University or Ohio State University to determine which has the third-largest university campus in the United States. To accurately answer the query, more detailed data on the campus sizes or student populations of both universities is required.''}. 
At higher temperatures ($T=2.0$), these perturbations often result in garbled or nonsensical outputs containing code snippets, random tokens, and mixed languages. \textit{NER Replacement} perturbations prove effective at disrupting model performance, causing failures in entity recognition and relationship inference that lead to incomplete or incorrect answers.

\textbf{Temperature Effects on Output Quality.}
Temperature settings influence how models degrade under perturbation. At $T=2.0$, we observe severe output degradation characterized by nonsense and complete failure to address the query. In contrast, at $T=0.6$, models demonstrate greater robustness, though they still exhibit cautious response patterns (e.g., ``I cannot determine...''), partial answers, and occasional factual errors. This suggests that while lower temperatures improve stability, they do not eliminate sensitivity to perturbations.

\textbf{Question-Type Sensitivity.}
Bridge questions show particular sensitivity to entity removal or replacement, likely due to their reliance on connecting information across multiple facts. Comparison questions, while still affected, occasionally maintain correctness through external knowledge utilization, suggesting different reasoning pathways may exhibit varying robustness.

\textbf{Model-Specific Degradation Patterns.}
The two models exhibit distinct failure characteristics. \texttt{gpt-4o} typically produces fluent but incorrect responses under perturbation, maintaining coherence while sacrificing accuracy. \texttt{deepseek-reasoner}, conversely, often fails more gracefully with concise but incomplete answers (e.g., responding with single words like ``Brewery'' rather than generating nonsensical text). 
This difference likely stems from their distinct training objectives; as a reasoning model, \texttt{deepseek-reasoner} may prioritize logical coherence and conciseness over the discursive fluency characteristic of a general-purpose model like \texttt{gpt-4o}, a hypothesis that merits further investigation.

\textbf{Robustness Insights.}
Despite overall sensitivity patterns, we observe instances where models maintain correctness under perturbation, indicating some degree of inherent robustness or effective internal knowledge utilization. The significant performance variability across samples suggests that certain question structures or knowledge domains are inherently more fragile than others. 

\section{Discussion and Conclusion}
\label{sec:discussion_conclusion}

This study investigated the relative impact of internal temperature versus external perturbations on RAG system performance. Our analysis reveals that temperature introduces a more pronounced influence on model correctness than specific perturbation types, with performance degrading significantly above certain temperature thresholds across most tested models and perturbation conditions.

The interaction between temperature and perturbations proves particularly critical: while models demonstrate relative robustness to perturbations at lower temperatures ($T\leq0.6$), they exhibit severe performance degradation under the same perturbations at higher temperatures ($T\geq1.4$). This joint effect creates a fragility landscape where systems that appear stable under standard evaluation conditions can fail dramatically when facing real-world noise combined with typical sampling strategies.

Notably, we observed instances where models maintained correctness despite substantial perturbations, suggesting utilization of internal knowledge rather than strict reliance on retrieved documents. However, the unpredictable nature of this phenomenon, where models sometimes bypass corrupted context entirely but other times produce confidently wrong responses, highlights the challenge of determining when and how internal knowledge mechanisms activate in RAG settings.

Our findings carry implications for RAG deployment. From a temperature perspective, we demonstrate that this hyperparameter must be carefully calibrated alongside perturbation robustness considerations. From a retrieval perspective, our results reinforce the importance of filtering uncertain or irrelevant content, aligning with principles in active retrieval methods like FLARE~\citep{jiang-etal-2023-active}. Based on our comprehensive evaluation, we propose the following deployment strategies: \texttt{deepseek-reasoner} for applications requiring consistent performance across diverse temperature settings; configure GPT models with a temperature ceiling of $T\leq1.4$ to avoid sharp performance degradation; and employ Llama models with a conservative temperature limit of $T\leq0.6$ to maintain acceptable correctness levels.

The main contribution of this work is to reveal the critical yet overlooked interaction between internal and external noise sources in RAG systems. A system that performs well on conventional benchmarks may prove surprisingly fragile when facing the combined effects of sampling stochasticity and real-world document perturbations. To address this gap, we introduce a dedicated benchmark and analytical framework designed to quantify this joint effect.
We note that the present study isolates the LLM generator's sensitivity by perturbing gold contexts, thereby controlling for retrieval noise. A future direction is to incorporate actual retrieval systems to examine how retrieval inaccuracies and generation sensitivity compound in end-to-end pipelines.

\section*{Limitations}
Our empirical findings are currently based on the HotpotQA dataset, which is a multi-hop factoid QA benchmark. While this allows for controlled perturbation analysis, the generalizability of the observed temperature-perturbation interaction patterns to other task types, such as summarization or open-domain dialogue, remains to be validated and is an important direction for future work.

This study focused specifically on temperature as the key stochasticity parameter, holding other generation parameters (e.g., top-p, frequency penalty) at default values. We acknowledge that these parameters could interact with retrieval noise, and their joint effects constitute a promising avenue for extending this framework.

\section*{Acknowledgments}
This work was partially funded by the ``Intelligent Systems for Bridging Data, Knowledge and Humans'' axis of the Grenoble Computer Science Laboratory (LIG). It was also conducted within the framework of the AugmentIA Chair, led by Didier Schwab and hosted by the Grenoble INP Foundation, thanks to the patronage of the Artelia Group. The chair also receives support from the French government, managed by the National Research Agency (ANR) under the France 2030 program with reference number ANR-23-IACL-0006 (MIAI Cluster).
We also thank anonymous reviewers for their insightful comments.

\section{Bibliographical References}\label{sec:reference}

\bibliographystyle{lrec2026-natbib}
\bibliography{lrec2026-example}

\section{Language Resource References}
\label{lr:ref}
\bibliographystylelanguageresource{lrec2026-natbib}
\bibliographylanguageresource{languageresource}

\appendix

\begin{table*}[!ht]
\centering
\small
\resizebox{\textwidth}{!}{%
\begin{tabular}{p{4cm}|p{10cm}}
\toprule
\textbf{Field} & \textbf{Content} \\
\midrule
Question & New Faces of 1952 is a musical revue with songs and comedy skits, it helped jump start the career of which young performer, and American actress? \\
Answer & Carol Lawrence \\
\midrule
\multicolumn{2}{c}{\textbf{Supporting Facts}} \\
\midrule
New Faces of 1952 (sent 2) & It helped jump start the careers of several young performers including Paul Lynde, Alice Ghostley, Eartha Kitt, Robert Clary, Carol Lawrence, Ronny Graham, performer/writer Mel Brooks (as Melvin Brooks), and lyricist Sheldon Harnick. \\
Carol Lawrence (sent 0) & Carol Lawrence (born September 5, 1932) is an American actress, most often associated with musical theatre, but who has also appeared extensively on television. \\
\midrule
\multicolumn{2}{c}{\textbf{Example Distractor Context}} \\
\midrule
Guess Who I Saw Today (sent 0) & "Guess Who I Saw Today" is a popular jazz song written by Murray Grand with lyrics by Elisse Boyd. \\
Guess Who I Saw Today (sent 1) & The song was originally composed for Leonard Sillman's Broadway musical revue "New Faces of 1952" in which it was sung by June Carroll. \\
\midrule
Monotonous (sent 0) & "Monotonous" is a popular song written by June Carroll and Arthur Siegel for Leonard Sillman's Broadway revue "New Faces of 1952". \\
Monotonous (sent 1) & The song was written based on the experiences of its singer Eartha Kitt. \\
\bottomrule
\end{tabular}
}
\caption{A HotpotQA bridge question example (ID: 5a76a401554299373536012b) from our evaluation set, showing supporting facts and sample distractor context.}
\label{tab:hotpotqa_example_bridge}
\end{table*}

\section{Dataset and Example}
\label{sec:appendix_dataset}

HotpotQA is distributed under a CC BY-SA 4.0 License. The dataset can be downloaded from: \url{https://hotpotqa.github.io/}. 
Table~\ref{tab:hotpotqa_example_bridge} presents an example from the HotpotQA dataset.

Our generated model outputs and experimental data are available on Hugging Face at: \url{https://huggingface.co/datasets/yongxin2020/TempPerturb-Eval-data}.

\end{document}

%% file: tables/perturbations_taxonomy.tex
\begin{tabularx}{\textwidth}{l l l l X}
\toprule
\textbf{Target} & \textbf{Category} & \textbf{Specific Method} & \textbf{Granularity} & \textbf{Description and Application} \\
\midrule

\multirow{4}{*}{\textbf{Document}} & \multirow{2}{*}{Subset} & Removal & Sentence-level & Identifies a minimal subset of sentences whose removal lowers the document's rank beyond a threshold \textit{k}. \\ 
& & & & \textit{Application:} Document ranking \citep{Rorseth2023CREDENCECE}, QA \citep{10.1145/3626772.3657660} \\ 
\cmidrule(lr){3-5}
& & Combination & Sentence/Passage-level & Identifies how combinations of elements influence results, often via fixed-size random sampling. \\ 
& & & & \textit{Application:} Open-book QA \citep{rorseth2024ragemachineretrievalaugmentedllm} \\ 
\cmidrule(lr){2-5}

& \multirow{2}{*}{Permutation} & Source Reordering & Passage-level & Identifies the effect of source order by finding permutations that place relevant sources in high-attention positions. \\ 
& & & & \textit{Application:} Open-book QA \citep{rorseth2024ragemachineretrievalaugmentedllm} \\ 
\cmidrule(lr){3-5}
& & Word Reordering & Word-level & Alters the sequence of words within each source of the input text. \\ 
& & & & \textit{Application:} Discussed in QA context \citep{10.1145/3626772.3657660} \\ 
\cmidrule(lr){2-5}

& \multirow{5}{*}{Replacement} & Unit Replacement & Sentence/Passage-level & Replaces one sentence or passage at a time. \\ 
& & & & \textit{Application:} Document ranking \citep{10.1145/3397271.3401058} \\ 
\cmidrule(lr){3-5}
& & Entity Replacement & Word-level & Identifies entities (nouns, proper nouns) and replaces them with random words. \\ 
& & & & \textit{Application:} Discussed in QA context \citep{10.1145/3626772.3657660} \\ 
\cmidrule(lr){3-5}
& & Antonym Replacement & Word-level & Replaces one or more words with their antonyms. \\ 
& & & & \textit{Application:} Discussed in QA context \citep{10.1145/3626772.3657660} \\ 
\cmidrule(lr){3-5}
& & Synonym Replacement & Word-level & Replaces one or more (important) words with their synonyms. \\ 
& & & & \textit{Application:} QA \citep{10.1145/3626772.3657660}, Document ranking \citep{10.1145/3576923} \\ 
\cmidrule(lr){2-5}

& \multirow{2}{*}{Injection} & Random Noise & Word-level & Inserts random words into or around the source text. \\ 
& & & & \textit{Application:} Discussed in QA context \citep{10.1145/3626772.3657660}, Passage ranking \citep{raval2020wordtimeadversarialattacks} \\ 
\midrule

\multirow{2}{*}{\textbf{Query}} & \multirow{2}{*}{Addition} & Prefix Injection & Token-level & Insertion of a short prefix to the prompt leads to generation of factually incorrect outputs. \\ 
& & & & \textit{Application:} QA \citep{10.1145/3637528.3671932} \\ 
\cmidrule(lr){3-5}
& & Term Augmentation & Token-level & Minimal perturbations to a search query that raise the rank of a given document. \\ 
& & & & \textit{Application:} Document ranking \citep{Rorseth2023CREDENCECE} \\ 
\bottomrule
\end{tabularx}